\documentclass[sigconf]{aamas} 

\usepackage{balance} 
\usepackage{amsmath}
\usepackage{graphicx}

\usepackage[utf8]{inputenc} 
\usepackage[T1]{fontenc}    
\usepackage{hyperref}       
\usepackage{url}            
\usepackage{booktabs}       
\usepackage{amsfonts}       
\usepackage{nicefrac}       
\usepackage{microtype}      
\usepackage{graphicx}
\usepackage{color}
\usepackage{caption}
\usepackage{subcaption}
\usepackage[ruled,vlined]{algorithm2e}
\usepackage{algorithmic}
\usepackage{algorithm2e}

\usepackage{float}
\usepackage{xcolor}
\usepackage{siunitx}
\usepackage[symbol]{footmisc}

\copyrightyear{2021}
\acmYear{2021}
\acmDOI{}
\acmPrice{}
\acmISBN{}

\title[Continuous Control with Deep Reinforcement Learning]{Continuous Control with Deep Reinforcement Learning for Autonomous Vessels
}

 \author{Nader Zare *}
 \affiliation{
   \department{Institute for Big Data Analytics}
   \institution{Dalhousie University}
   \city{Halifax}}
 \email{nader.zare@dal.ca}

 \author{Bruno Brandoli *}
 \affiliation{
   \department{Institute for Big Data Analytics}
   \institution{Dalhousie University}
   \city{Halifax}}
 \email{brunobrandoli@dal.ca}

 \author{Mahtab Sarvmaili}
 \affiliation{
   \department{Institute for Big Data Analytics}
   \institution{Dalhousie University}
   \city{Halifax}}
 \email{mahtab.sarvmaili@dal.ca}

 \author{Amilcar Soares}
 \affiliation{
   \department{Department of Computer Science}
   \institution{Memorial University of Newfoundland}
   \city{St. John's}}
 \email{amilcarsj@mun.ca}

 \author{Stan Matwin *}
 \affiliation{
   \department{Institute for Big Data Analytics}
   \institution{Dalhousie University}
   \city{Halifax}}
 \affiliation{
   \department{Institute for Computer Science}
   \institution{Polish Academy of Sciences}
   \city{Warsaw}}
 \email{stan@cs.dal.ca}

\begin{abstract}

Maritime autonomous transportation has played a crucial role in the globalization of the world economy.
Deep Reinforcement Learning (DRL) has been applied to automatic path planning to simulate vessel collision avoidance situations in open seas. 
End-to-end approaches that learn complex mappings directly from the input have poor generalization to reach the targets in different environments.
In this work, we present a new strategy called state-action rotation to improve agent's  performance in unseen situations by rotating the obtained experience (state-action-state) and preserving them in the replay buffer. 
We designed our model based on Deep Deterministic Policy Gradient, local view maker, and planner.
Our agent uses two deep Convolutional Neural Networks to estimate the policy and action-value functions.
The proposed model was exhaustively trained and tested in maritime scenarios with real maps from cities such as Montreal and Halifax. Experimental results show that the state-action rotation on top of the CVN consistently improves the rate of arrival to a destination (RATD) by up 11.96\% with respect to the Vessel Navigator with Planner and Local View (VNPLV), as well as it achieves superior performance in unseen mappings by up 30.82\%. Our proposed approach exhibits advantages in terms of robustness when tested in a new environment, supporting the idea that generalization can be achieved by using state-action rotation. 

\end{abstract}


\newcommand{\BibTeX}{\rm B\kern-.05em{\sc i\kern-.025em b}\kern-.08em\TeX}

\begin{document}


\pagestyle{fancy}
\fancyhead{}

\fancyhf{}
\rfoot{\vspace{1mm}\thepage}


\maketitle 

\sloppy
\section{Introduction}
\label{sec:introduction}

Deep Reinforcement Learning (DRL) has become one of the most active research areas in Artificial Intelligence (AI). This remarkable surge started by obtaining human-level performance in Atari games \cite{mnih2015human} without any prior knowledge, and continued by defeating the best human player in the game of Go \cite{silver2016mastering}. Since then DRL has been vastly employed for robotic control \cite{gu2017deep,johannink2019residual,tai2017virtual}, strategic games \cite{jaderberg2019human,silver2017mastering,silver2018general}, and autonomous car navigation \cite{sallab2017deep,isele2018navigating}. Thereby, DRL has a strong potential to process and make decisions in challenging and dynamic environments, such as autonomous vessel navigation and collision avoidance. 

Traditionally, ship navigation has been performed entirely by humans. However, the advancement of maritime technologies and AI algorithms have big potential to make the vessel navigation semi or fully autonomous. An autonomous navigation system has the ability to guide its operator in discovering the near-optimum trajectory for ship collision avoidance \cite{statheros2008autonomous}. This idea is known as path planning, in which a system attempts to find the optimal path in reaching the target point while avoiding the obstacles. The proposed algorithms in \cite{di2015energy,duchovn2014path,mac2016heuristic} require complete environmental information in the real world. This unknown environment is explored by a single or multiple agents, that is, prior knowledge for training the agent is a restricting assumption. In addition, to find the shortest path from a departure point and a landing location is posed as an challenge. That is because in autonomous navigation vessels cannot navigate close to the shore, usually areas considered shallow and entirely with stones and terrain banks. Regarding RL-based navigation, the main concern is about model generalization for different locations.

This problem was addressed by A* algorithms \cite{cheng2014} and Q-learning \cite{Magalhaes2018}.
In this context, Yoo and Kim\cite{Yoo2016} proposed a new reinforcement learning method for path planning of maritime vessels in ocean environments; however, without considering obstacles. Q-Learning is very limited to low-dimensional state spaces domains. Unlike, in \cite{cheng2018} developed a DRL-based algorithm for unmanned ships based on DQN with no prior knowledge of the environment. Similar to our approach, they used the local view strategy to reduce the state-space exploration and path planning more efficiently.
In contrast with Cheng and Zhang, we designed our model based on Deep Deterministic Policy Gradient. The main drawback proposed by the authors is regarding the discrete actions for reaching the target. Recently, \cite{shen2019} presented a prototype of multiple autonomous vessels controlled by deep q-learning. The main issue with this idea is the fact they incorporated navigation rules employed with navigational limitation by polygons or lines.

This research's main goal is not to find the shortest path for an autonomous vessel, but to develop a new strategy of continuous action on top of the local view path planning. Our proposal is designed based on the DDPG architecture by using the environment's raw input images to learn the best path between a source and a destination point. The idea is to feed the network with the input images through the actor to predict the best continuous action for that state, and then to feed the predicted action and the state to the critic network to predict the state-action value. The critic network output is used during the training phase, and when this phase is over, the actor will select actions. Our model was exhaustively trained and tested in maritime scenarios with real data maps from cities such as Montreal and Halifax with non-fixed origin and destination positions. For training the model, we used areas with the presence of real geographical lands. 
We give a detailed example about state-action rotation in Section \ref{sec:sar}. This process is trained by using the DDPG alorithm. We chose DDPG because it holds an actor-critic architecture. We describe the DDPG model in Section \ref{sec:background}. The process is started over with the module transition maker with the reward given the state of the environment. The agent exploration converges when either target is reached or he hit obstacles in the scenario, such as rocks and islands along the vessel's navigability.


In summary, the main contributions of our paper are:
\begin{itemize}
	\item The development of an environment for 2D simulation of real-world maps that takes insights of the Floyd and Ramer-Douglas-Peucker algorithms to find the trajectory between two positions.
	\item We developed a new architecture named Continuous Vessel Navigator (CVN) based on the DDPG algorithm with continuous action spaces, evaluating generalization in two different shores.
	\item Our proposal uses the environments as raw images. We emphasize that the training is performed with raw images or maps, however, this could be extended for a real-scenario with cameras on the unmanned vehicles.
\end{itemize}

The rest of this paper is organized as follows.
In Section \ref{sec:background}, we described two architectures of RL that are used in this work. 
Section \ref{sec:sar} details our proposed rotation strategy (SAR) using a motivating example. In Section \ref{sec:basicelem}, we introduce the basic elements of reinforcement learning for an ocean environment, and a new model called Continuous Vessel Navigator (CVN) to navigate a ship from an origin to a destination autonomously. 
Section \ref{sec:exps} demonstrates the efficiency of SAR and CVN, and it discusses the results of the experiments in a simulated maritime environment. 
Finally, the conclusions of this work and future directions come in Section \ref{sec:conclusion}.

\section{Background}
\label{sec:background}

\subsection{Deep Q-Network}

\emph{Deep Q-Network (DQN)} proposed by Mnih et al. \cite{mnih2015human} is a model-free RL algorithm that uses a multi-layered neural network $Q$ to approximate $Q^*$. It has been exploited in continuous or high dimensional environments with a discrete action space. DQN receives state $s$ and outputs a vector of action values $Q(s,a)$. However, the RL models are unstable when a non-linear function approximator is exploited to estimate the Q-values. There are several reasons for this problem: the correlations between sequences of observations that result in a significant change in agent policy after each update and also the correlation between action-values and target values. A randomized replay buffer is used to solve the correlations between the observation sequences. 
To address the second issue, another neural network known as a target network with $\theta ^ -$ parameters has been exploited to decorrelate the relation between the action-values and target-values. Hence, in DQN there are two neural networks; the online network that uses $\theta$ as weights, and the target network that uses $\theta^-$.

During training, the environment sends state $s_t$, and a reward to the agent and the agent selects an action by a $\epsilon-$greedy policy; in the next step, the agent generates transition tuples $(s_t, a_t, r_t, s_{t+1})$ and stores the tuples in the replay buffer. The $\epsilon$-greedy policy of $Q$ selects a random action with probability $1 - \epsilon$ and $\pi_Q(s)=argmax_{a \in A}Q(s,a)$ with probability $\epsilon$. To train the online network, we select a batch of transitions from the replay buffer and train the online network by using mini-batch gradient descent using the Bellman equation (Equation \ref{eq:3}). 

\begin{equation}
\label{eq:3}
\begin{split}
\ell=\mathbf{E}(Q(s_t,a_t) - y_t)^2 \ \textrm{where,} \\
Y^{DQN}_t\equiv R_{t+1}+\gamma \max_{a}Q(S_{t+1},a;\theta^{-}_{t})
\end{split}
\end{equation}

After each $\tau$ steps, we copy the online neural networks weights $\theta$ to target neural network $\theta^-$.\\

\subsection{Deep Deterministic Policy Gradient}
\label{sec:DDPG}

\emph{Deep Deterministic Policy Gradient (DDPG)} \cite{lillicrap2015continuous} is a model-free RL algorithm that uses two multi-layered neural networks for training the agent in continuous action environments. The first net is called the actor, which estimates the policy of the agent $\pi: S \xrightarrow[]{} A$; the second network is called the critic, which is responsible for estimating the action-value function $Q: S \times A \xrightarrow \ {\rm I\!R}$. 
During training, the environment sends state $s_t$ and a reward to the agent, which selects an action by a noisy version of the target policy. Afterward, the agent generates the transition tuples $(s_t, a_t, r_t, s_{t+1})$ and stores the tuples in the replay buffer similarly to DQN. The noisy version of the target policy uses $\pi_b(s)=\pi(s)+\mathcal{N}(0,1)$. 

The critic network is trained similarly to the Q-function presented by DQN; however, we use the outputs of the actor network to predict the Q-value of next state according the Formula \ref{eq:4}.
Basically, several approaches in reinforcement learning are based on the recursive Bellman equation:

\begin{equation}
\label{eq:4}    
Q(s,a): y_t=r_t+\gamma Q(s_{t+1},\pi(s_{t+1}))
\end{equation}

The actor neural network is trained with gradient descent; the loss function employed to train the actor is shown in Equation \ref{eq:5}.

\begin{equation}
\label{eq:5}
\mathcal{L}_a=-{\rm I\!E}Q(s, \pi(s))
\end{equation}

\section{Path Finding with State-Action Rotation}
\label{sec:sar}
In this section, we use a simple environment created to verify the effect of the State-Action Rotation (SAR) strategy on the baseline DQN architecture. 
Positive outcomes of the SAR algorithm led us to apply this method on the path planning problems.
In this section, we present the idea of State-Action Rotation strategy in Section \ref{sub:rotation}, and then we test its impact in a simple environment, described in Section \ref{sub:simple_tests}. Positive outcomes of the SAR algorithm led us to apply this method on the path planning problems.

\subsection{State-Action Rotation Strategy}
\label{sub:rotation}
\begin{figure}[h]
  \centering
  \includegraphics[width=0.48\textwidth]{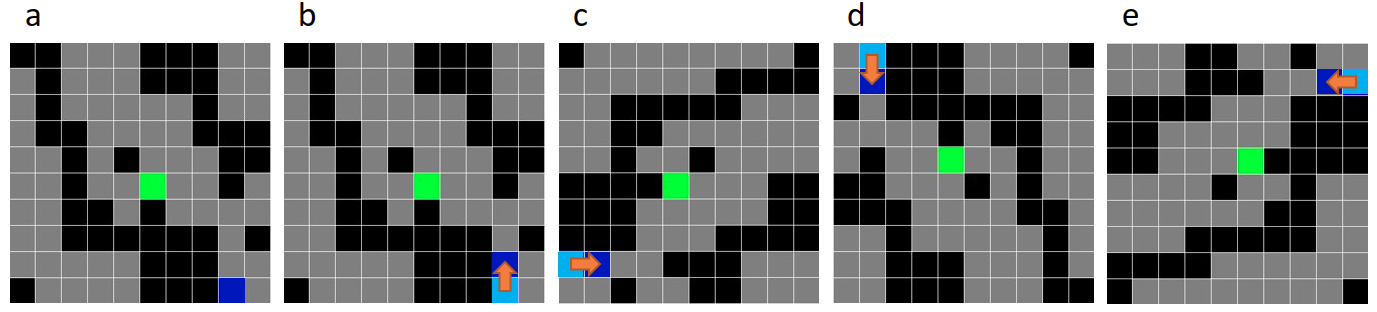}  
\caption{\label{fig:rotationSample}Visualization of path finder environment and rotation function on the environment. (a) path finder environment, (b) original $s_t, a_t, s_{t+1}$, (c) 90$^{\circ}$, (d) 180$^{\circ}$ and (e) 270$^{\circ}$ rotated transitions.}
\end{figure}

The idea behind \textit{SAR} is simple: after experiencing each step, the agent generates rotated transitions by using the next state $s_{t+1}$, current state $s_t$, and selected action $a_t$, and stores them in the replay buffer. 
We specify the number of rotations with respect to the environment (the number of possible actions in that environment); for instance, in the path planning problems, the agent rotates each transition 3 times, and it finally stores 4 transitions in its replay buffer.
To better explain this process, we visualize how agent rotates the obtained experience of time step $t$. 
Figure \ref{fig:rotationSample} illustrates the result of rotating a transition.
At the time step $t$ the agent is located at $s_t$ (dark blue square) and it wants to move towards the goal (green square) Figure. (a).\ref{fig:rotationSample}. At this time step, it selects the $a_t$ (the red arrow) and moves toward the next state $s_{t+1}$ (dark blue). Now agent creates a tuple of the original $s_t$ (light blue square), $a_t$ (red arrow), $s_{t+1}$ (dark blue square), and then it rotates them with 90$^{\circ}$, 180$^{\circ}$, 270$^{\circ}$ degrees to create three more tuples. Finally agent preserves all of the obtained experiences in the replay buffer.

There is at least one path from the agent's position to the target's position in all of the generated maps. The agent's position is shown with a blue pixel and the target's position with a green pixel. The pseudo-code of the rotation algorithm is given in Algorithm \ref{alg:1}. In order to adopt the SAR algorithm, one does not need to change the structure of RL algorithm.

\begin{algorithm}[!h]
    \SetAlgoLined
    \textbf{\emph{Input:}} an off-policy RL algorithm $\mathbb{A}$, \\
    \hspace{1cm} a reward function $r: S \times A \Longrightarrow \mathbb{R}$ \\
    \textbf{\emph{Output:}} Hyperparameter configuration, trained agent and value functions \\    
    Initialize $\mathbb{A}$ \\
    Initialize replay buffer R \\
    \textbf{\emph{for:}} episode = 1, M \textbf{\emph{do:}} \\
        \quad Initial observation state $s_0$ \\
        \quad \textbf{\emph{for:}} t = 0, T - 1 \textbf{\emph{do:}} \\
            \quad \quad sample an action $a_t$ using the behavioral policy from $\mathbb{A}$\\
            \quad \quad \quad $a_t \Leftarrow \pi(s_t)$\\
            \quad \quad Execute the action $a_t$ and observe a new state $s_{t+1}$\\
        \quad \textbf{\emph{end for}} \\
        \quad \textbf{\emph{for:}} t = 0, T - 1 \textbf{\emph{do:}} \\
            \quad \quad $transition_t = (s_t, s_{t+1}, r_t, a_t)$\\
            \quad\quad Store $transition_t$ in R \quad \textrm{standard experience replay} \\
            \quad \quad \textbf{\emph{for:}} rot = 1, ROT \textbf{\emph{do:}} \\
                \quad \quad \quad $transition_t=rotate(trainsition_t)$ \\
                \quad \quad \quad Store $transition_t$ in R\quad SAR\\
            \quad \quad \textbf{\emph{end for}} \\
        \quad \textbf{\emph{end for}} \\
        \quad \textbf{\emph{for:}} t = 1, N \textbf{\emph{do:}} \\
            \quad \quad Sample a minibatch B from the replay buffer R\\
            \quad \quad Preform one step of optimization $\mathbb{A}$ and minibatch B \\
        \quad \textbf{\emph{end for}} \\
    \textbf{\emph{end for}} \\
    \caption{State-Action Rotation} 
    \label{alg:1}
\end{algorithm}

\subsection{A simple environment and early analysis of the SAR}
\label{sub:simple_tests}

\begin{figure}[htbp]
\begin{subfigure}{0.11\textwidth}
  \includegraphics[width=1\textwidth]{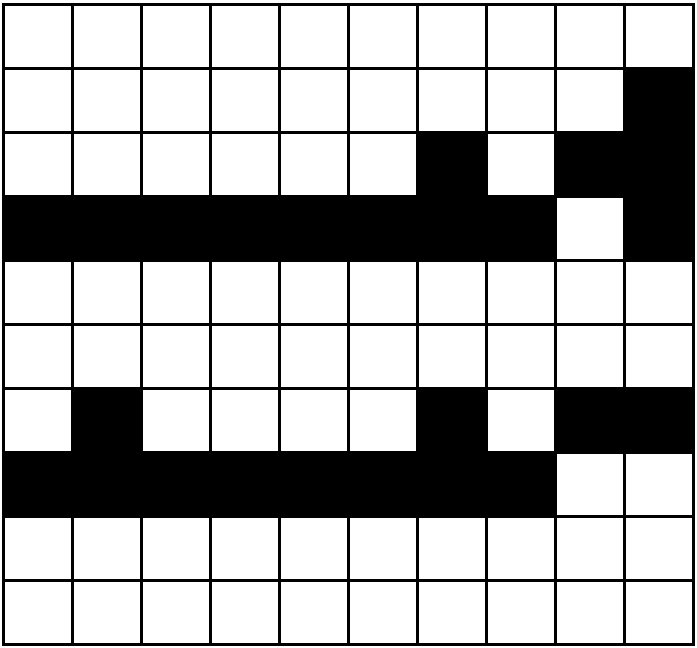}  
  \caption{}
\end{subfigure}
\hfill
\begin{subfigure}{0.11\textwidth}
  \centering
  \includegraphics[width=1\textwidth]{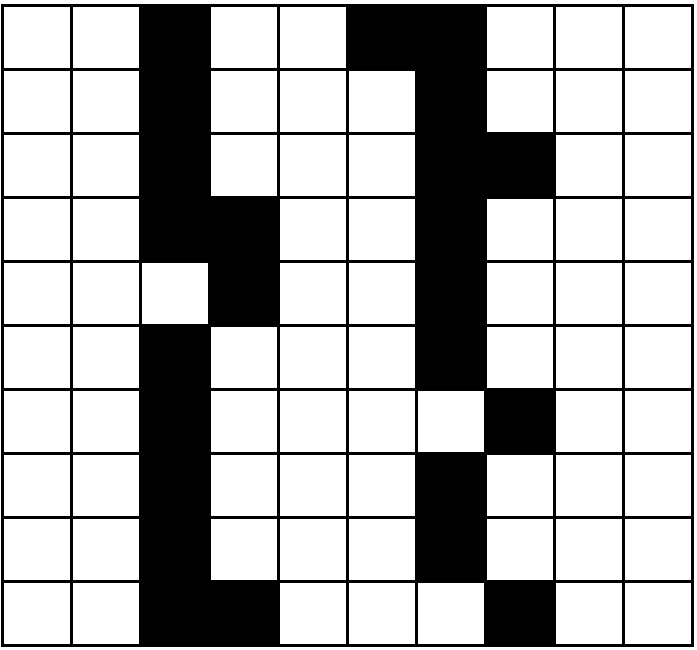}  
  \caption{}
\end{subfigure}
\hfill
\begin{subfigure}{0.11\textwidth}
  \centering
  \includegraphics[width=1\textwidth]{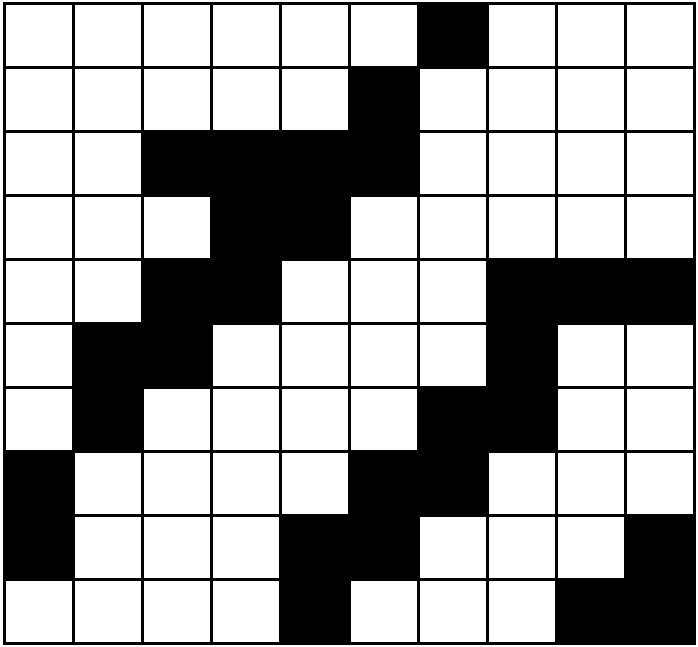}  
  \caption{}
\end{subfigure}
\caption{\label{fig:rotationMapExample}An example of three rendered maps, (a) a horizontal map, (b) a vertical map and in (c) a diagonal map.}
\end{figure}

We implemented a simple environment, a $10 \times 10$ pixel image, that contains walls, a target, an agent, and some free pixels that an agent can travel within. The agent should reach the target position without hitting the walls or going outside of the field. The action space of the agent contains $A = \{up, down, left, right\}$. At each episode, the environment renders one of the horizontal, vertical or diagonal maps (see Figure \ref{fig:rotationMapExample}). In the beginning of the process, the environment generates a random position for the agent and a random position for the target.

In this motivational experiment we addressed the example of Figure \ref{fig:rotationMapExample} with discrete actions. The reward function sends back a reward of +1 to the agent in case it reaches the target, while -1 if it hits any wall or goes outside of the field. We calculated the distance between the agent position and the target based on the difference of the agent's previous and current positions by aiming at giving a small positive or negative reward to the agent. We limited each episode to a maximum number of fifty steps in the environment unless it terminates earlier. 

For this example and throughout this paper, we adopted the Rate of Arrival To Destination (RATD) metric, which is widely adopted accuracy measure to calculate the performance of the agent, which refers to the percentage of successful plans $P$ in $N$ randomly generated plans, i.e., $RATD = 100 * (|P| / |N|)$. 

\begin{figure*}[ht]
\centering
\includegraphics[width=1\textwidth,trim=50 190 350 0, clip]{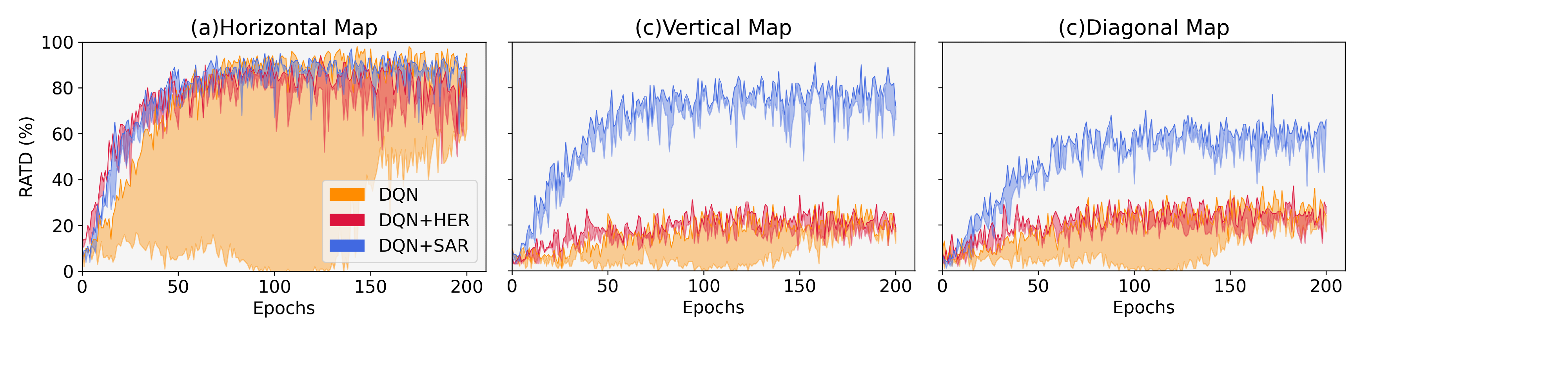}
\caption{Results using the average RATD for the algorithms DQN, DQN with HER, and DQN with SAR tested on three mappings: (a) horizontal (b) vertical and (c) the diagonal maps.}
\label{fig:rotationResults}
\end{figure*}

We then tested different architectures, specifically DQN \cite{mnih2015human}, DQN with HER \cite{andrychowicz2017HER} (DQN + HER), and DQN with our state-action rotation strategy (DQN + SAR). The idea is to understand the outputs according to the the number of epochs versus the performance RATD measure, displayed in the plots of Figure \ref{fig:rotationResults}. We trained an agent in horizontal maps only, and then tested it in horizontal, diagonal, and vertical maps. 
200 epochs were set for training the agents, and each epoch contains 1,000 training episodes and 100 testing episodes for those three types of maps. The agent that adopted the SAR strategy attained better performance in vertical and diagonal maps. Our idea is to confirm by comparing the results that the usage of rotation strategy can impact positively on the performance of a certain agent in unseen situations.
It is important to mention that, once the environment does not support continuous action, we cannot use actor-critic architectures, such as DDPG (see Section \ref{sec:DDPG}).


\section{Basic Elements of Reinforcement Learning Model}
\label{sec:basicelem}


In this section, we define the environment used in our maritime scenario simulation (Section \ref{sec:environment}) and discuss the modifications done in widely adopted algorithms for path-planning, such as the Floyd and Rammer-Douglas-Peucker algorithms (Section \ref{sec:planner}). After, in Section \ref{sub:cvn}, we present our proposed model - Continuous Vessel Navigator - with Planner and Local View (CVN).


\subsection{Environment definitions}
\label{sec:environment}

In this section, we detail the environment exploited by our unmanned vessel simulator.

\noindent\textbf{Agent.} In the Unmanned Vessel Simulator, an agent is a vessel, and its goal is voyaging safely from an origin position through the water without hitting any land or any obstacles and arriving at a destination circle by using the sequence of continuous or discrete actions. 

\noindent\textbf{Map.} It is the environment is an area that contains a body of water with presence of terrain and small islands.

\noindent\textbf{Origin Point.} The origin point is initial position that the agent starts at the first moment of each training or testing phase ($t = 0$). The origin point is a pair $OP = (x_s, y_s)$, where $x_s$ is the agent's latitude at the first step of training or testing phase, and $y_s$ is the longitude of the agent at that time. The environment randomly selects the origin point for each phase.

\noindent\textbf{Destination Circle.} The destination circle is a round area that the agent should voyage towards and reach. The destination circle is shown by a triplet $DC=(x_d, y_d, r_d)$, where $x_d$ and $y_d$ are the longitude and latitude of the center position of the destination circle (destination position) and $r_d$ is the radius from the center point (the destination radius). The environment randomly selects the center position of the destination circle for each phase, such that there is at least one way from the origin point to the destination circle without considering obstacles.

\noindent\textbf{Intermediate Goal.} It is the goal of the agent in each episode that the agent should reach. It comes in the state from the environment to the agent. The idea is a circle and is shown by a triplet $IG=(x_g, y_g, r_g)$, where $x_g$ and $y_g$ are the latitude and longitude of the referred goal and $r_d$ is the radius of this goal that is equal to the destination circle radius.

\noindent\textbf{Episode Start Point.} The \textit{episode start point} is the position of the agent at the first step of a new episode, and it is equal to the last position of the agent in the previous episode in a phase. The episode start point is a tuple $SP=(x_e, y_e)$, where $x_e$ and $y_e$ are the latitude and longitude of the position of the agent.

\noindent\textbf{Action.} We used two types of actions in the Unmanned Vessel Simulator, continuous action, and discrete action. In the continuous action mode, the agent defines a tuple $(v_x, v_y)$ where $v_x$ is the momentum of the agent in the latitude, and $v_y$ is the momentum of the agent in the longitude of the environment. In this paper, the agent's maximum velocity in the latitude and the longitude is 0.001 degrees. In the discrete action mode the agent can select an action from the action set $A=\{N,S,E,W,NE,NW,SE,SW\}$. The agent's position is changed 0.0005 or 0.001 degrees in latitude or longitude by using a discrete action in this work.

\noindent\textbf{State.} State is the local view with observation of the agent from the environment.
Each pixel in the local view can be land, water, obstacle, or target. The vessel position is always in the center of the state. In this work, the environment makes the local view of the agent and sends it as a state to the agent.

\noindent\textbf{Outcomes.} The possible outcomes of the actions can be summarized as follows: \textit{O} = \{hit an obstacle, hit land, arrive at the target, vanish target, normal movement\}. \textit{\textbf{hitting an obstacle}} or \textit{\textbf{hitting the land}} happens when the agent hits one of the vessels moving in the environment, or the agent took a direction that directs it to a geographic land area respectively. \textit{\textbf{vanish target}} happens when the intermediate goal has been removed from the agent's local view. If the agent successfully reaches its intermediate goal, the outcome of the action is \textit{\textbf{arrive at target}}. Finally, when the mission of the agent has not finished yet, the output is \textit{\textbf{normal movement}}. 

\noindent\textbf{Step.} The action of the agent and its outcome in the environment is considered as a step $s_i=(a_i,o_i)$, where $s_i \in S, a_i \in A, o_i \in O$, and S is the set of all possible steps. Each step moves our agent from the current state to the next state.

\noindent\textbf{Episode.} A set of consecutive steps with a fixed start point and an intermediate goal is identified as an \textit{episode}. In each episode, the agent starts from the episode start point with the objective of arriving at its goal. The episode is defined as $e_j = (SP_j, IG_j, <s_1, s_2,\cdot, s_n >)$ where the $SP_j$ is the start point and $IG_j$ is the agent's intermediate goal, and $<s_1, s_2,\cdot, s_n >$ is the sequence of steps in that episode. $e_j \in E$ is the set of all possible episodes. The outcome of the last step in each episode is considered as the final outcome of the episode.

\noindent\textbf{Plan.} It is defined with an origin point $OP_k$, a destination point $DP_k$, and the set of episodes $<e_1, e_2, \cdot, e_m>$. The intermediate goals are the destinations in each episode $(e_1, e_2, \cdot, e_m)$  of a plan $p_k$. However, the target of the last episode is same as the final Target. The final outcome of a plan determines the success or failure of that plan. So, if the outcome of the final episode in the plan is not reaching the target, that plan is considered to be a \textbf{\textit{failed plan}}. On the other hand, if the agent (i) has arrived at target and (ii) the agent is in its destination, then the plan is identified as a \textbf{\textit{successful plan}}.

\subsection{Path-planner}
\label{sec:planner}

 The idea to employ the planner aims to decrease the number of states and, consequently, improve the learning process's efficiency. We divided the path between the origin point and the destination point to shorter paths. Thus, the planner helps the agent to voyage a long distance with better performance. In the first step, the planner finds the shortest path from the origin point to the destination point using the Floyd algorithm \cite{floyd1962algorithm}. The planner saves the shortest path between each pair of points for each map to improve the performance. In the second step, the environment uses the Rammer-Douglas-Peucker (RDP) algorithm\cite{hershberger1994n} or Land Attended Rammer-Douglas-Peucker to decrease the number of intermediate points in this path. We use RDP to give flexibility to the agent in making local decisions. 
 
\begin{figure}[h]
    \centering 
    \includegraphics[width=8cm]{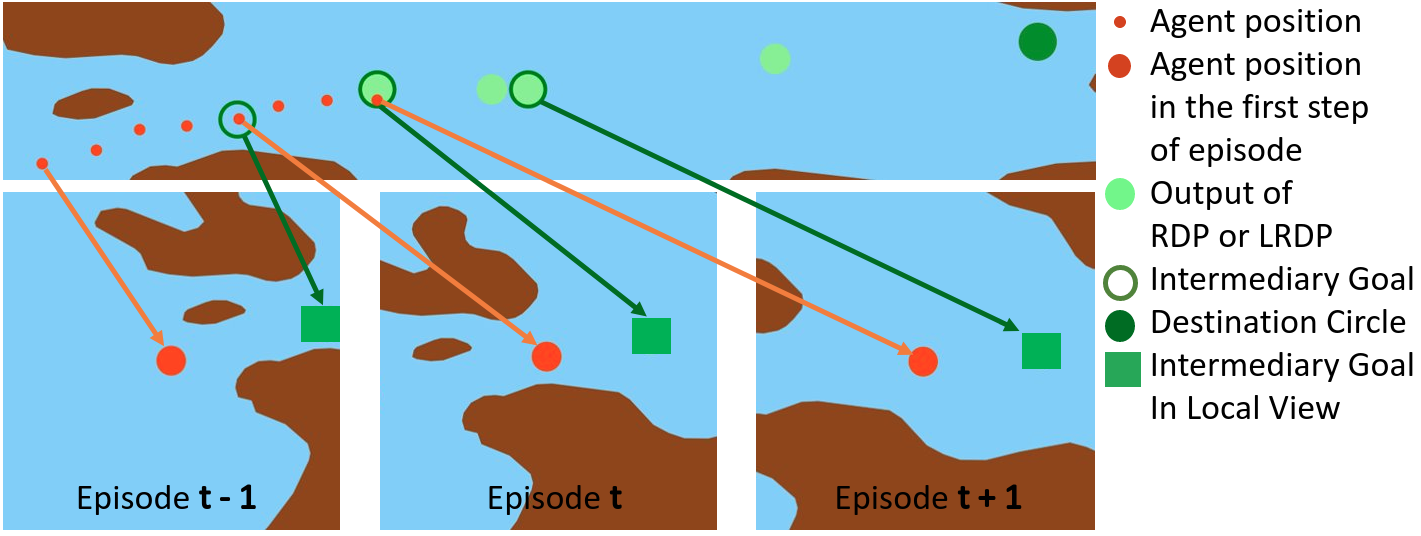}
    \caption{\label{fig:PLANNER}Example showing how to select the intermediate goal of the plans based on the output of the Rammer-Douglas-Peucker (RDP) algorithm.
    }
\end{figure}

 Figure \ref{fig:PLANNER} illustrates the functioning of the Path Planner. The output of the RDP algorithm is a list of points that have been recursively selected from the original set of points. The planner then specifies a pseudo target that is in the local view of the agent at the beginning of each episode. This point is an intermediate goal that the agent attempts to reach in each episode. The planner defines a new intermediate goal in the agent's local view after the agent arrives at the current intermediate goal or at the beginning of the new plan. In the first step of each episode, we determine the intermediate goal for the local view; hence we look at the outputs of RDP. In this case, three situations may occur: first, only one point of the $points \in RDP$ exists in the local view boundary, so that point is considered as the goal. Secondly, if two points of $points \in RDP$ exist in the area of local view, the furthest point is chosen as the goal. In the last case, none of the $points \in RDP$ are located inside the local view, so we draw a line from the start position of the agent to the next closest point, and the intersection of this line with the boundary of the local view is selected as the intermediate goal.
 
\subsubsection{Modified Floyd}
\label{sec:Modifiedfloyd}

\begin{figure}[h]
    \centering
    \subfloat[Floyd]{{\includegraphics[scale=0.075]{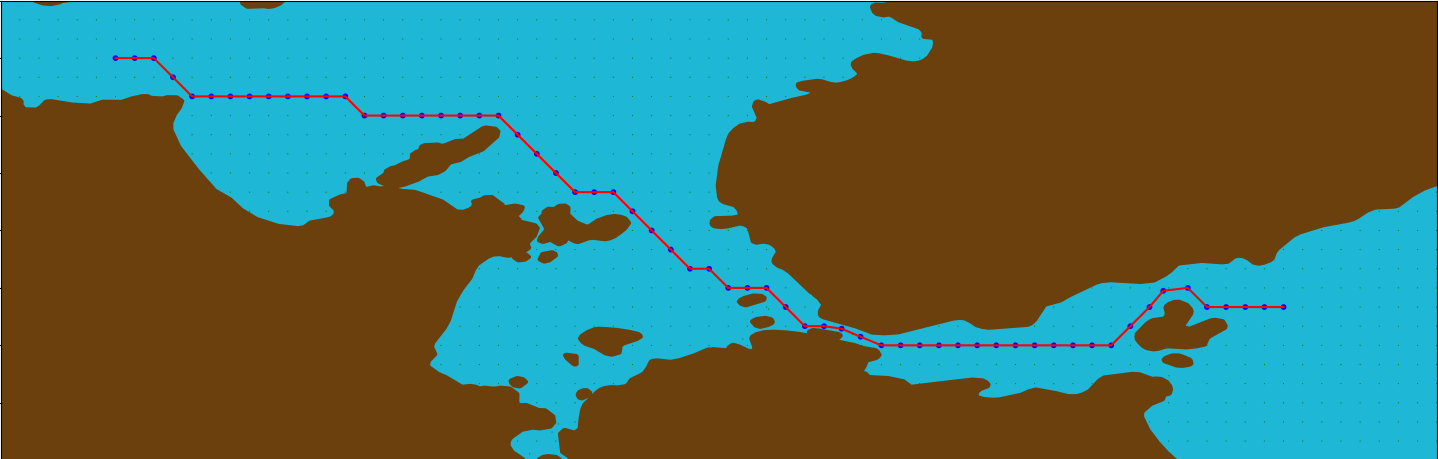}}}
    \qquad
    \subfloat[Modified Floyd]{{\includegraphics[scale=0.075]{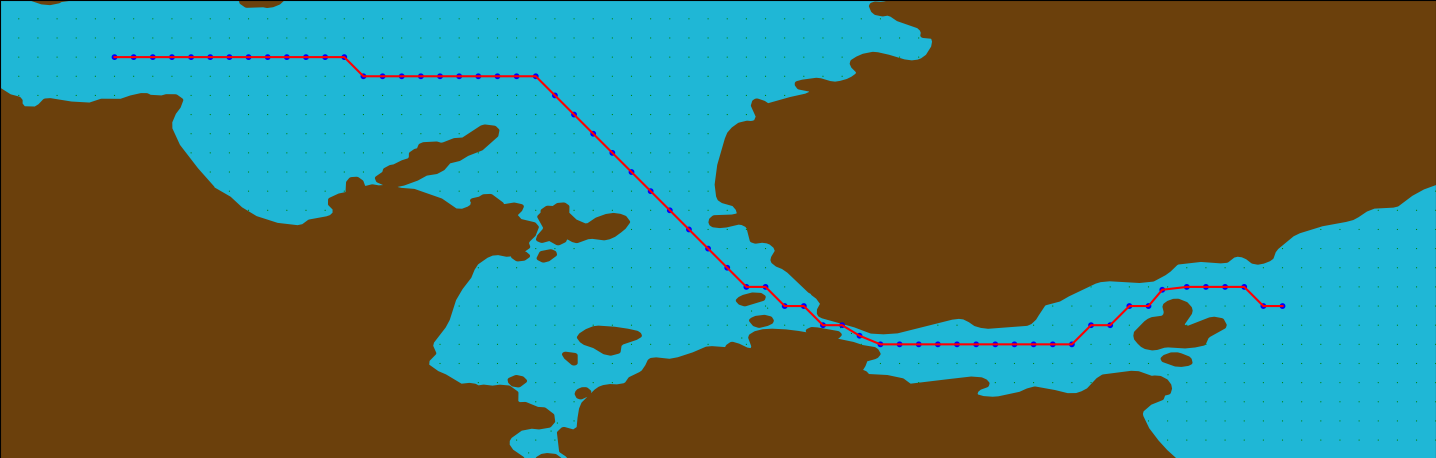}}}
    \caption{\label{fig:safe}The visualization of two trajectories (in red color) from source point to destination point by using Floyd and Modified-Floyd algorithms, respectively.} 
\end{figure}

The original Floyd algorithm has been employed to find the shortest path in a graph with positive or negative weights\cite{floyd1962algorithm}. However, this algorithm cannot find the shortest path with negative cycles. In this paper, we tested our models with two types of weight assignments to find the shortest path between origin and destination points. In the first type, we set the distance of each pair of neighbors according to the weight of the edges in the graph. In the second type, we measure the weights according to the distance of each point and its neighbours to the nearest land. To find the weight between points $A$ and $B$, we count the number of neighbours that are located in the land ($l$) and in the water ($w$) for point $B$, then we count the neighbours in land ($lan$) and in water ($wan$) for each neighbour of point $B$. Finally, we calculate the weight using the following formulation. We called this algorithm as Modified Floyd. The output between Floyd and Modified Floyd in Montreal map is shown in Figure \ref{fig:safe}. Modified Floyd attempts to find safer and distant path from the terrain banks.


\subsubsection{Land Attended Rammer-Douglas-Peucker}

The traditional Rammer-Douglas-Peuker (RDP) algorithm basically reduces the number of points in a trajectory using a distance threshold\cite{hershberger1994n}. RDP receives a trajectory represented as a point set $PL = {P_0, \ldots, P_{end}}$, where $P_0$ is the start point and $P_{end}$ is the end point.
The algorithm then selects a point with maximum distance $P_{max}$ between two farthest points. To this end, the algorithm selects the start and the ending point. Subsequently, the idea is to split the trajectory into new smaller trajectories. 
If the distance of this point to the line segment is smaller than the distance threshold, other points are discarded. On the other hand, if the distance of the point farthest from the line segment is greater than the distance threshold, then that point should be kept. The algorithm is recursively applied to both trajectory subparts to the new start and the farthest points. The algorithm stops when the trajectory subparts are retained if the distance $d_{max}$ is lower than the threshold. This idea is widely used for data trajectory compression.


Nonetheless, the regular RDP does not consider obstacles (e.g. land areas), 
so the agent cannot find the best action to reach to intermediate goal. In our environment, if the value of the threshold is considered small, the agent cannot be flexible in terms of decision-making to select an action; otherwise, if the value of the threshold is large, it may select points such that the shortest path between them places into the land.

We introduce a modified version of the $RDP$ algorithm that is called Land Attended Rammer-Douglas-Peucker (LARDP). Our proposed algorithm considers an additional condition to remove a candidate point from the list of points $PL$, so that we keep the candidate point if it is either larger than the threshold or if the segment between previous and next points is placed on the land. The pseudo-code of LARDP is showed in Algorithm \ref{LAR}.

\begin{algorithm}[!htbp]
    \SetAlgoLined
    \textbf{\emph{function LARDP}(PL[], threshold):} \\
        \quad \textbf{\emph{if }}\textbf{size}(PL) == 1:\\
            \quad \quad \textbf{\emph{return }}PL[0]\\
        \quad $d_{max},i_{max}$=\\
        \quad\quad\quad\textbf{max}$_i$ (\textbf{distance}((PL[$0$],PL[$i$]),(PL[$i$],PL[$end$])))\\
        \quad on\_field = \textbf{IsOnField}(\textbf{line}(PL[$0$],PL[$end$]))\\
        \quad \textbf{\emph{if }}$d_{max}$ > threshold \textbf{\emph{or }} on\_field:\\
            \quad \quad PL1 = \textbf{LARDP}(PL[0:$i_{max}$], threshold)\\
            \quad \quad PL2 = \textbf{LARDP}(PL[$i_{max}$:$end$], threshold)\\
            \quad \quad \textbf{\emph{return}} PL1[0:$end$) + PL[$i_{max}$] + PL2(0:$end$]\\
        \quad \textbf{\emph{else: }}\\
            \quad \quad \textbf{\emph{return}} PL[0] + PL[$end$]\\
    \caption{Land Attended Rammer-Douglas-Peucker}
    \label{LAR}
\end{algorithm}

\subsubsection{Local View Maker}
\label{sec:localviewmaker}
\begin{figure}[H]
    \centering 
    \includegraphics[width=0.5\textwidth]{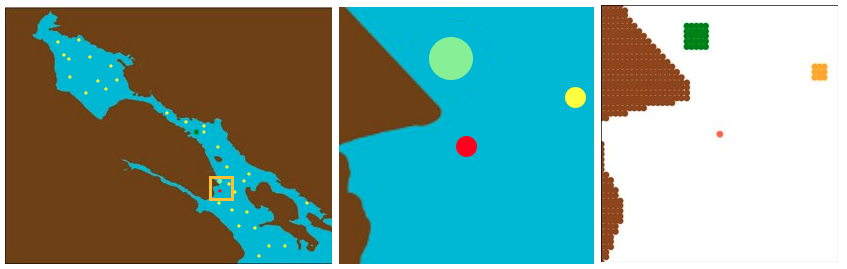}
    \caption{\label{fig:Localview}An screenshot of the simulation performed with the local view strategy over the Halifax map. The agent is depicted in red color, while green color stands for the intermediate goal, and yellow color shapes display the obstacles.}
\end{figure}

The local view maker is a part of the environment that receives the map, the position of the obstacles, the intermediate goal position, and the agent's position. Afterwards, it forms the state as a $51 \times 51$ pixel image to feed any specific convolutional neural network. In this case, we show in Figure \ref{fig:Localview} an example that simulates the same real environment of the Halifax map. We randomly selected 30 obstacles from AIS data, here considered as obstacles, but their positions are changed in each plan.

\subsection{Continuous Vessel Navigator}
\label{sub:cvn}

The Continuous Vessel Navigator (CVN) is an algorithm for training autonomous vessels in dynamic environments. 
CVN uses raw input images of the environment to learn the best path between a source and a destination point. 
The algorithm is designed based on the DDPG method; hence it processes the input images through the actor and predicts the best (continuous) action for that state. 
After, CVN feeds the predicted action and the state to the critic network to forecast the state-action value. 
The critic network output is used during the training phase, and when this phase is over, the actor selects the best action.

\begin{figure}[h]
    \centering 
    \includegraphics[width=0.45\textwidth]{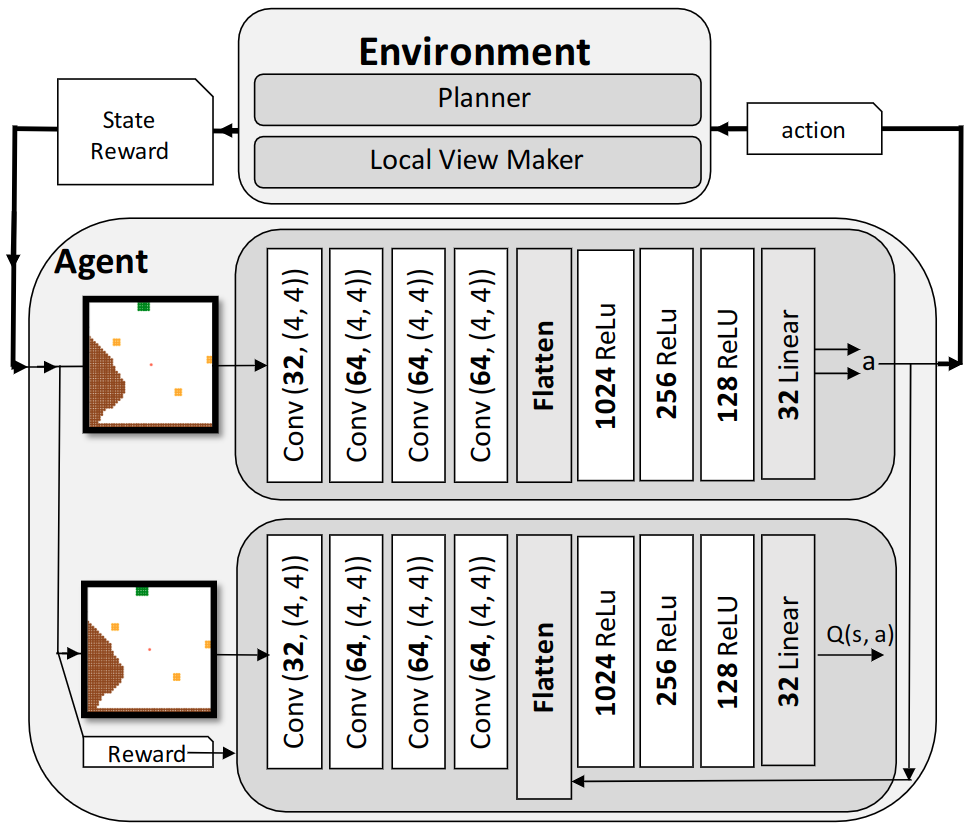}
    \caption{\label{fig:VNPLV}Proposed architecture named Continuous Vessel Navigator with planner and localview (CVN).}
\end{figure}

\section{Experiments}
\label{sec:exps}

\subsection{Baseline}


Here, we compared our proposed CVN against the Vessel Navigator with Planner and Local View (VNPLV) \cite{etemad2020using}, which uses DQN architecture. VNPLV is a method for autonomous vessel navigation in large scale environments that takes advantage of the path planner and local view approach. It is split into 3 steps: (i) making a full plan from the start point to the destination point by using the Floyd Algorithm; (ii) reducing intermediate-similar goals by applying the Ramer-Douglas-Peucker algorithm;  and (iii) deciding on the details of the plan such as arriving at the shadow of the intermediate goal in the local view of the agent. 

The environment sends the local view as state and reward to the agent for both VNPLV and CVN.
In such way, the agent that uses VNPLV method just contains a Deep Convolution Neural Network. 


The CVN can be seen as an extended version of VNPLV.
However, unlike VNPLV that deals with discrete actions, CVN was designed to deal with continuous actions. 
The CVN uses two separate Convolutional Neural Networks (CNN), one as a policy network (Actor), and one as a value estimator (Critic) to estimate the action-value function. 
The actor network is responsible for learning the optimum policy, and the critic network estimates the value of selected action in the given state \cite{sutton2018reinforcement}.


\subsection{Dataset}
\label{sub:dataset}

We selected two regions to test our model on the Canadian shore: Halifax and Montreal. In Halifax we selected the bounding box (\emph{longitude}, \emph{latitude}) starting from ($-63^\circ 69 ^\prime, 44^\circ58^\prime$) and ending at ($-63^\circ49^\prime, 44^\circ73^\prime$), and in Montreal we selected a bounding box starting from ($-74^\circ10^\prime,45^\circ30^\prime$) and ending at ($-73^\circ70^\prime,45^\circ50^\prime$). We trained our set of agents with the Halifax map and tested them in both Halifax and Montreal. We used public data on earth elevation using the NOAA\footnote{https://coast.noaa.gov/dataviewer/} dataset in order to create a water and a land layer. 
We randomly selected 30 obstacles from Automatic Identification System (AIS) data for each plan; they remain constant during that plan, but dynamic to the other plans. 


\subsection{Training and testing setup}
\label{sub:setup}
In this work, we trained and tested six types of agents that are listed in Table \ref{tab:agents}. In our training phase, we trained each agent in 40 batches that contain 1,000 training plans and 100 testing plans. During the training phase, we stored transitions (state, reward, next state, and action) in the replay buffer \cite{mnih2015human} for stabilizing the learning process at the end of each step during the training of A1, B1, and C1. 

After each step of the training for the agents A2, B2, and C2, we employed the SAR strategy for storing experiences (transitions) in the replay buffer. It generated three new rotated transitions based on the original transition and stored them in the replay buffer. In our proposed framework, we took advantage of a replay buffer with a size of 100,000 transitions.
In the training phase of VNPLV agents, they select action using a $\epsilon \ greedy$ policy that is described in Formula \ref{eg:1}, where $pn$ is the number of the current plan in the training phase, so the weight of exploring decreases from 1.0 to 0.1 in the first half of the training phase and, in the second half, it remains as 0.1. However, the CVN agents use the Ornstein-Uhlenbeck process \cite{uhlenbeck1930theory,lillicrap2015continuous} to select noisy actions. In the testing phase, all six agents choose the best action using their policy. 

\begin{equation}
\small
\pi(x) = \begin{cases}
    \text{random action of A(s)}, & \epsilon < max(0.1, 1.0 - pn / 20,000 * 0.9) \\
    argmax_{a\in A(s)} Q(s,a), & \text{Other}\\
\end{cases}
\label{eg:1}
\end{equation}

The idea of the training set up of the CVN agents was inspired by the paper of Etemad et al. \cite{etemad2020using}. At each step, we retrieved 32 transitions from the replay buffer to train the networks and, then, after every 200 steps, we updated the target networks using the online networks' weights.
After each batch, we save the models' parameters, so after all batches, we have 40 models, then we find the best agent using RATD of each agent in the testing plans of each batch.
Finally, we tested each of the selected agents from the three test configurations at the testing phase. 
For the first test, we evaluated the models on the Halifax map; in the second test set, we evaluated our models on the Montreal map; finally, we assessed the models on the Montreal map, but we compared RDP, LRADP, and Modified Floyd.

\begin{table}[!htbp]
  \centering
  \begin{tabular}{clllc}\toprule
    \textit{agent name} & \textit{method} & \textit{action type} & \textit{velocity} & \textit{SAR}\\ 
    \midrule
    A1 & VNPLV & discrete & 0.001 & No \\
    B1 & VNPLV & discrete & 0.0005 & No \\
    C1 & CVN & continuous & 0.001 & No \\ 
    \hline
    A2 & VNPLV & discrete & 0.001 & Yes \\
    B2 & VNPLV & discrete & 0.0005 & Yes \\
    C2 & CVN & continuous & 0.001 & Yes \\
    \bottomrule
  \end{tabular}
  \caption{Setting of six cases represented by different agents and combination of parameters and SAR strategy.}
  \label{tab:agents}
\end{table}

\subsection{Assessment of the maximum distance between origin and destination points}
\label{sub:effect_max}

In this section, we experimentally evaluated how significant is the effect of distance between the origin and the destination points using the metric RATD. 
To evaluate this change performed by each agent,
we set the maximum distances (in degrees) for six different values - 0.01, 0.02, 0.04, 0.08, 0.16, and 0.32 - totalizing 36 configurations. 

We trained with each of these combinations, and then we selected the best agents with the best performance during the training phase.
In addition, during the training phase, we used the Modified Floyd and LARDP algorithms. 
We also tested the best agents for another 10,000 plans in Halifax. For the testing phase, we also used Modified Floyd and LARDP in the same way as in the training phase.

\begin{figure}[h]
    \centering
    \includegraphics[width=0.50\textwidth]{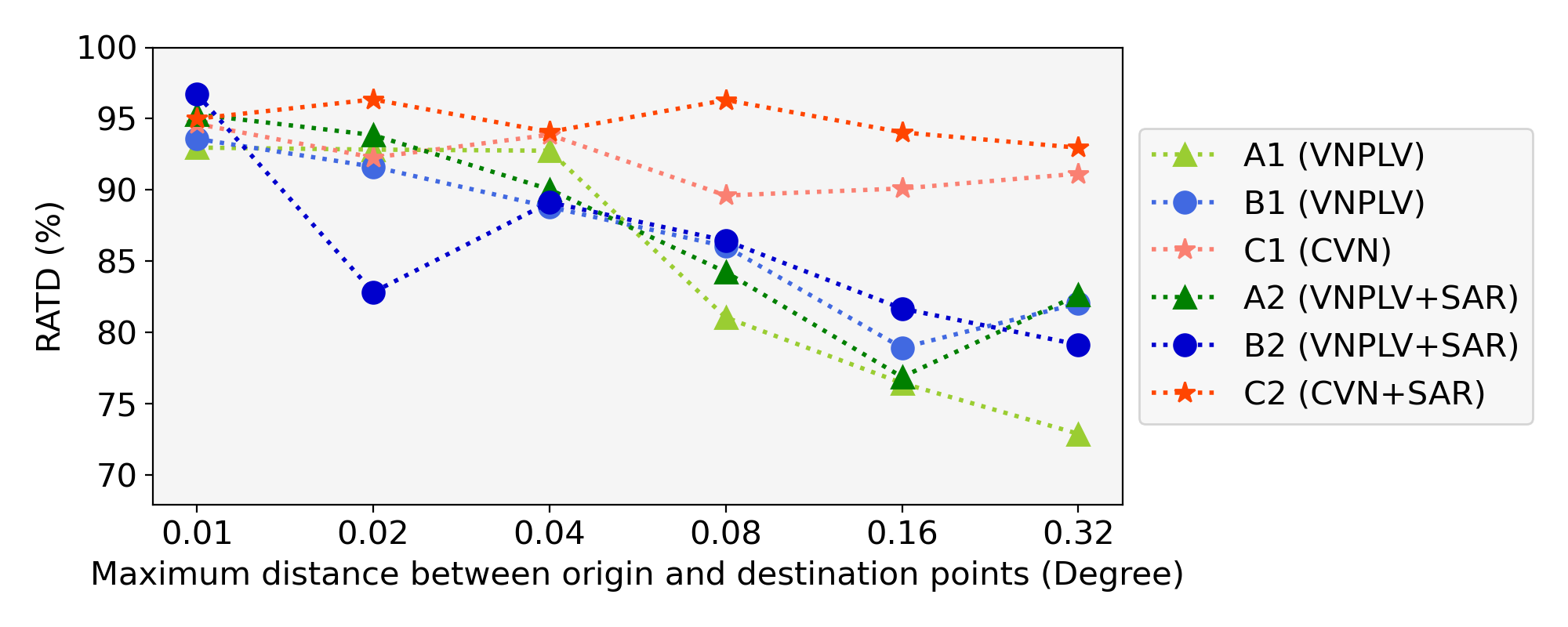}
    \caption{\label{fig:halifax_same}The plot shows if distance is  larger between the origin and destination point the RATD metric is reduced for the DQN network (VNPLV). Nevertheless, the DDPG-based CVN agents was not significantly affected by this change.}
\end{figure}

Figure \ref{fig:halifax_same} shows the RATD when the agents were trained and tested in the same environment configuration.
This plot reveals that the CVN agents are more successful in reaching the destinations than VNPLV agents when the distance between the origin point and the destination point increases. For the CVN agents, the C2 that uses the SAR strategy performs slightly better than the C1. 

\begin{figure}[htbp]
    \centering
    \includegraphics[width=0.50\textwidth]{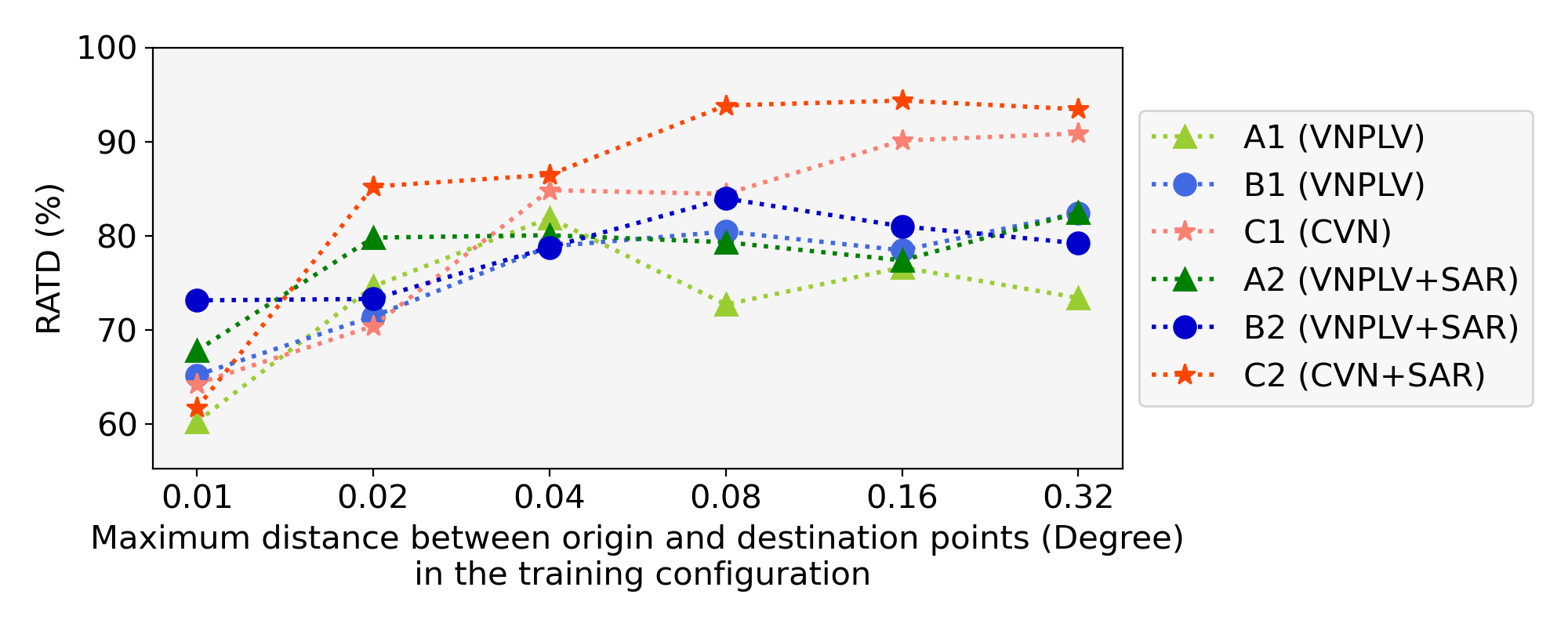}
    \caption{\label{fig:halifax_32}
    Assessment of the distance increment between source and destination points. The idea here is that training was performed in a configuration set using the same distance from the previous figure, but the the test environment had no constraint applied on the distance between the source and destination, that is, it was random set to perform the exploration.}
\end{figure}

In contrast to the previous experiment, Figure \ref{fig:halifax_32} shows the results of our experiment when we tested the agents in a different environment configuration after that the agent was trained. 
The plot shows that training agents with larger distance between the origin and destination points have better RATD in comparison with shorter distances for random placement of the start and end points. That is explained because once if we increase the distance, it leads the agent to explore different states of the environment when compared with shorter distances, and consequently, tends to achieve better result. In this case, we tested each agent on the Halifax map.

\begin{figure}[h]
    \centering
    \subfloat[$d_{max} = 0.01$]{{\includegraphics[width=0.15\textwidth]{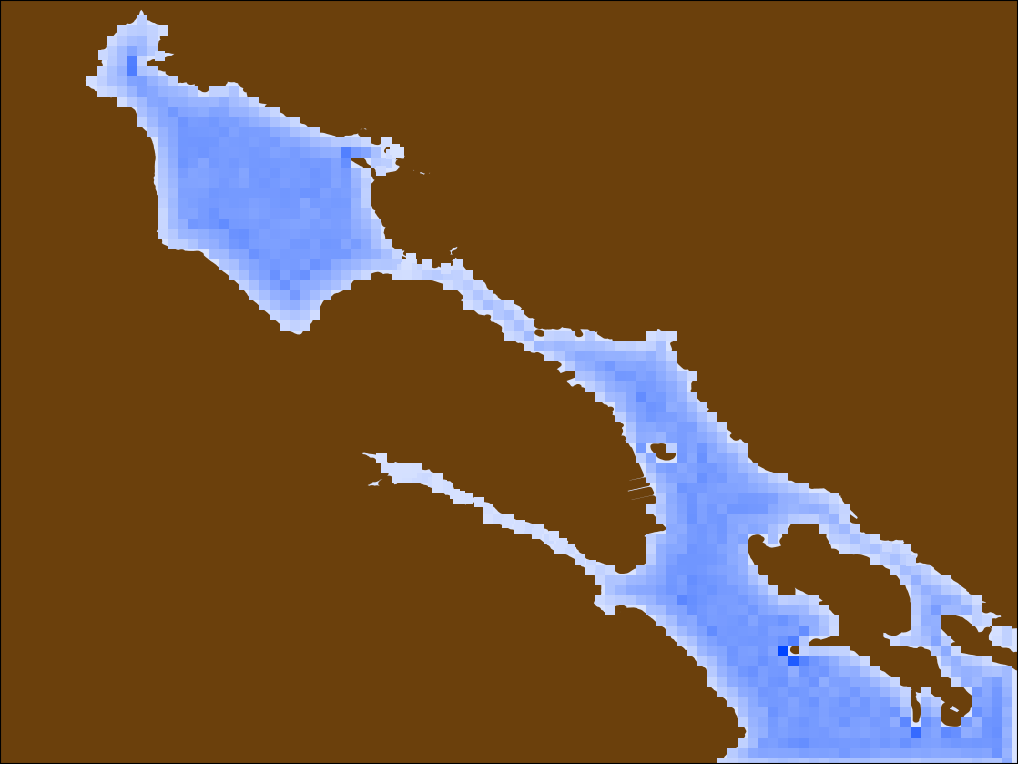}}}
    \hfill
    \subfloat[$d_{max} = 0.08$]{{\includegraphics[width=0.15\textwidth]{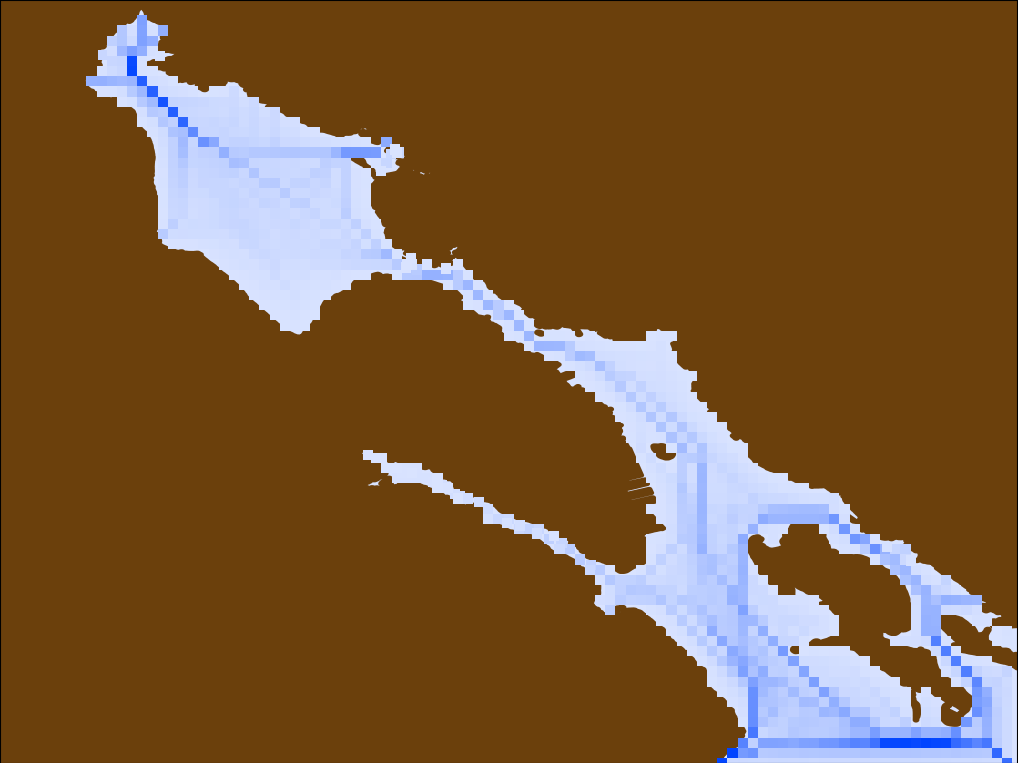}}}
    \hfill
    \subfloat[$d_{max} = 0.32$]{{\includegraphics[width=0.15\textwidth]{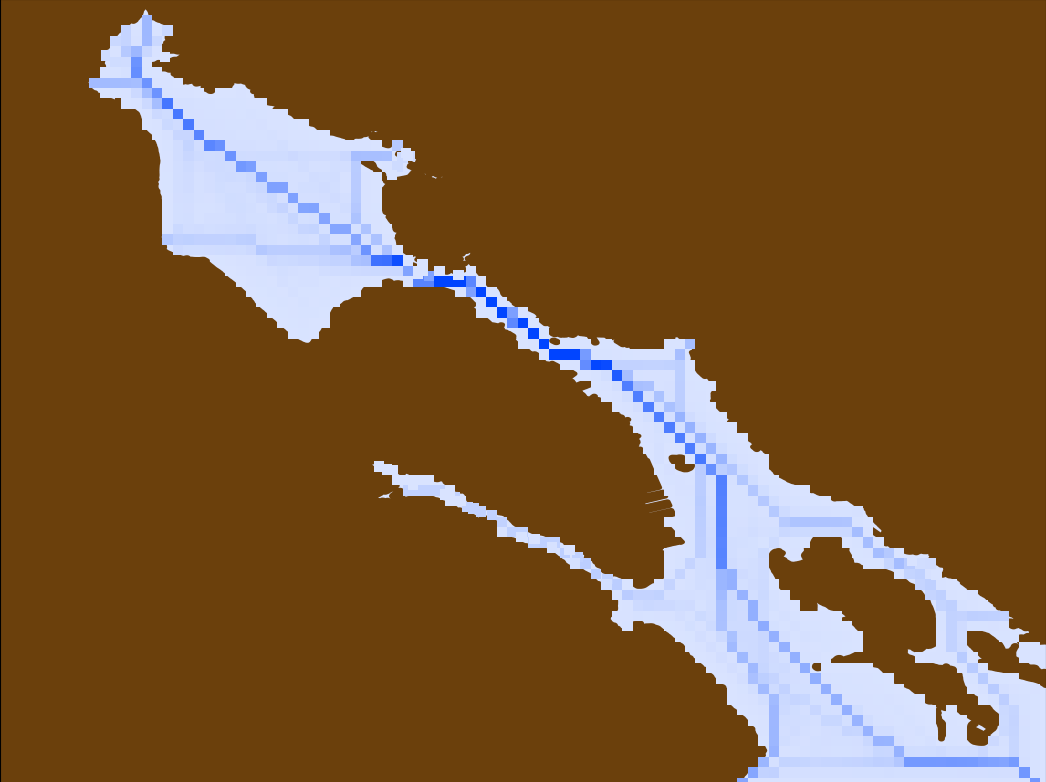}}}
    \caption{\label{fig:difdist}Evaluation of the maximum distance relating to the visited places on the map. Here we show the density visualization map of three different values for the maximum distance parameter.}
\end{figure}

The main explanation for the results achieved in the plot of Figure \ref{fig:halifax_32} is presented in Figure \ref{fig:difdist}. Results corroborate with the idea of the more you explore complex parts of the environment, such as narrow paths and straits, the more your model learns and tends to avoid collisions providing higher RATD. That is because such areas are the most critical for agents in terms of collision on vessel navigation maps; thus they need to be explored repeated times. 
In this experiment, we ran our position generator to yield randomly ten thousand tuples of origin and destination points, that is, with the dynamic origin and target positions. We drew the path between each of these origins and destination points using the Floyd path planner algorithm. We then counted the number of times that each pixel was visited in order to visualize the density exploration map. 
Figure \ref{fig:difdist} shows the visualization of each pixel visited when we set the maximum distance, between the origin and destination points, to the values 0.01, 0.08, and 0.32.
Therefore, we demonstrate that the larger is the distance, the greater change of agents to explore straits and maximize the ability to learn in such areas. 

\subsection{Generalization to a New Map}

\begin{figure}[h]
    \centering
    \includegraphics[width=0.49\textwidth, trim=50 50 0 50, clip]{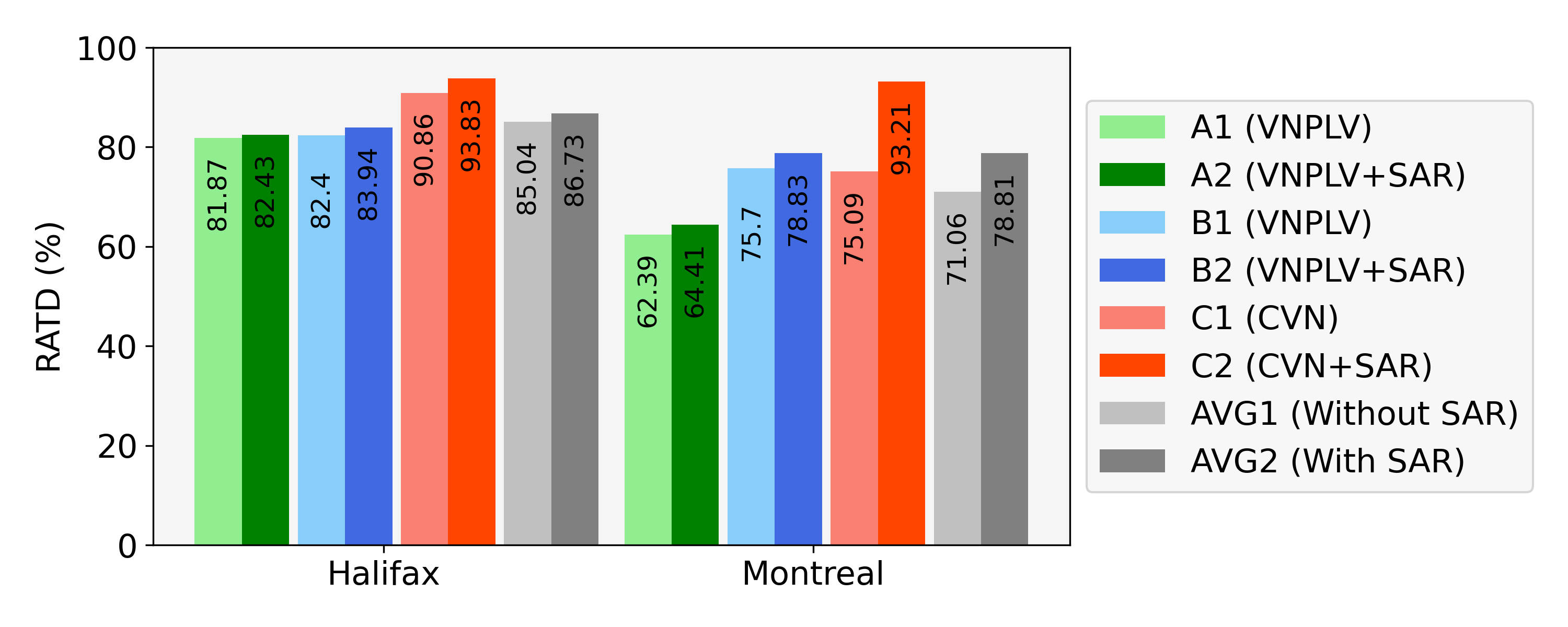}
    \caption{\label{fig:halifax_montreal}Results of the RATD metric for all agents trained in Halifax and tested in Montreal.} 
\end{figure}

In this section, we evaluate how is the generalization performance of our model on an unseen map. 
This step can be really challenging for the models since the agent has only explored different areas of a training environment. 
In this experiment, we selected the best agents from the six combinations (as seen in Table 1), and then we tested them on the Montreal map. 
The results show that agent C2 has the best performance on the Montreal map. 
Figure \ref{fig:halifax_montreal} illustrates the RATD for our test set. 
The same figure 
Figure \ref{fig:halifax_montreal} shows that the use of SAR algorithm, outperformed the results on both the Halifax map and the Montreal Map when compared with the other approaches without the rotation strategy. 
By using the SAR strategy, the results show that RATD increased from 0.56\% to 2.97\% and 1.69\% on average on the Halifax map, while on the Montreal map, the RATD metric increased from 2.02\% to 18.12\% and 7.75\% on average. Additionally, our proposal CVN with SAR has the largest difference, that is, CVN improves the results when compared to VNPLV based on DQN architecture.
It is important to point out that the Montreal map that we used to test our agents is 1.7 times larger than Halifax map.

\subsection{Improvement of the Modified Floyd and LARDP algorithms}

\begin{figure}[h]
    \centering
    \includegraphics[width=0.48\textwidth, trim=50 70 50 50, clip]{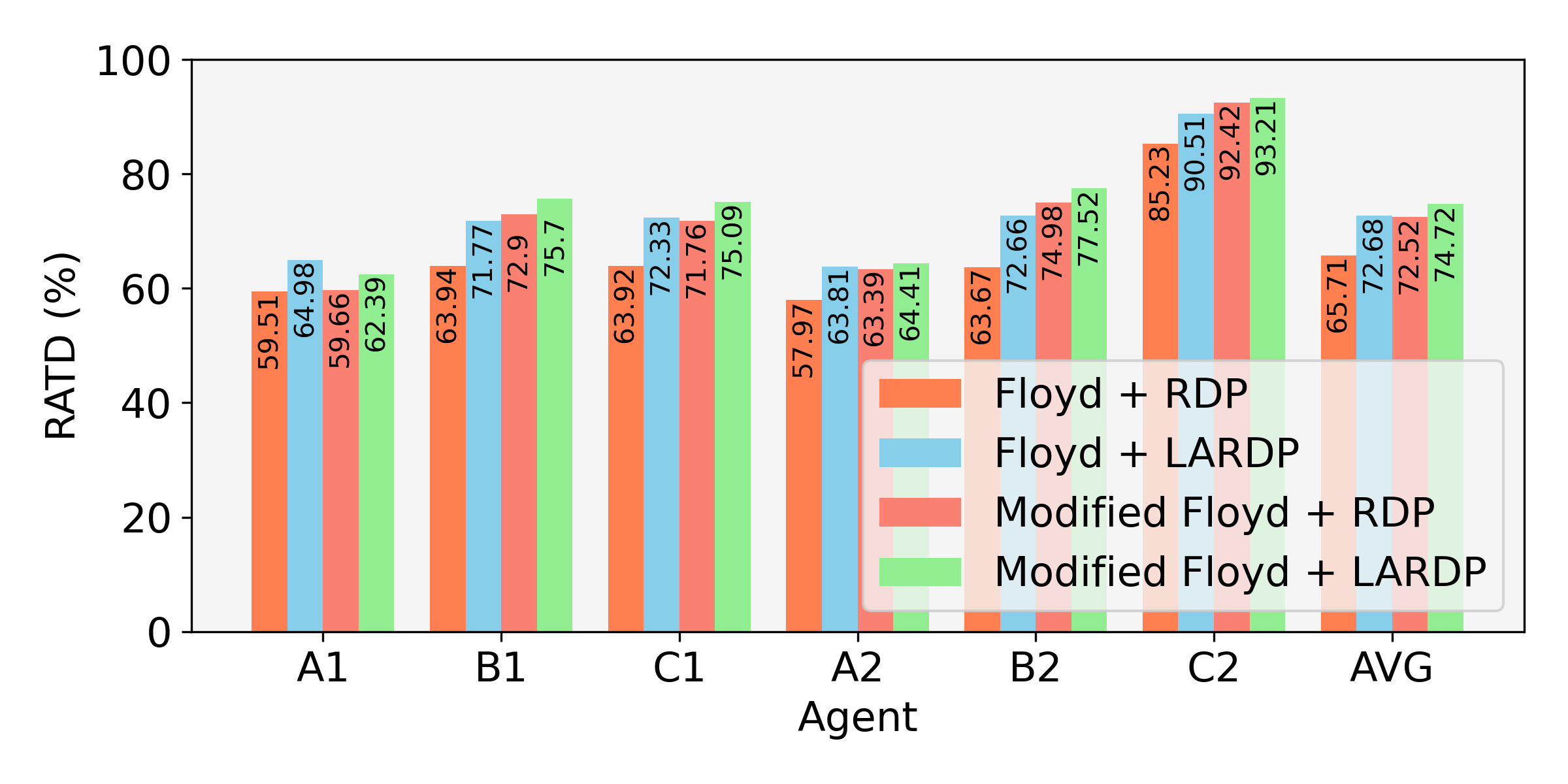}
    \caption{Results of the RATD metric with different planning methods over the Montreal map.}
    \label{fig:montreals}
\end{figure}

To demonstrate the effects of our new planning approach (i.e., a combination of Modified Floyd and Land Attended RDP) on the agent's performance,  we tested the best trained agents four times in the Montreal Map. In the first case of our experiments, we employed the Floyd+RDP planning method. In the second case, we used Floyd+LARDP as the planner for the agents. In the third case, we used Modified-Floyd+RDP, and finally, we exploited the Modified-Floyd+LARDP. Figure \ref{fig:montreals} shows the results of the experiments.
The results show that using LARDP improves RATD 4.58\% on average, and using Modified-Floyd increased RATD 4.42\% on average. As we expected, the RATD drops when we test the model in a different environment; however, the C2 agent was less affected, and its performance was nearly the same as the training agent.

\section{Conclusion}
\label{sec:conclusion}

We proposed a novel approach based on the Deep Deterministic Policy Gradient to perform automatic path planning of maritime vessels, which is an essential element for autonomous shipping and collision avoidance. 
Our model is a Continuous Vessel Navigator (CVN) that uses Convolutional Neural Networks (CNN) to process raw input images for decision-making. 
To this end, we presented an approach to generalize the learning of continuous control policy on real-world vessel navigation environment simulation. In the context of generalization, a new strategy named SAR strategy was developed to improve the agent's performance in unseen situations by rotating the obtained experience and preserving the new generated data in the replay buffer. 

In order to test our system, the simulation used a map of the Halifax harbor area. We validated the generalization of the proposed model by evaluating it in the Montreal harbor. The results indicated that the proposed algorithm based on the DRL could avoid ship collisions efficiently. Nonetheless, it can outperform results purely based on the local planner when SARS is adopted. Our experiments used simulation tasks to learn the policy while using real-world data to adapt the simulated model to real-world dynamics. The performance improvements were obtained not only on the open sea with static obstacles but also in severely congested and restricted waters. Further investigation in this area is a promising approach for real-world autonomous vessel learning, particularly for AIS-based autonomous vessels. 
As a future direction of this research is to compare our methodology with Hindsight Experience Replay (HER).

\bibliographystyle{ACM-Reference-Format} 
\bibliography{sample}

\end{document}